\documentclass[final,3p,times]{elsarticle}

\graphicspath{{Fig/}}
\usepackage{amssymb}
\usepackage{amsmath}

\usepackage{booktabs}
\usepackage{multirow}
\usepackage{dsfont}
\usepackage{threeparttable}
\usepackage{color}
\journal{Artificial Intelligence in Medicine}

\begin{document}

\begin{frontmatter}

\title{DRExplainer: Quantifiable Interpretability in Drug Response Prediction with Directed Graph Convolutional Network}


\author[ustc,ahau,equal]{Haoyuan Shi} 
\ead{haoyuan.shi@mail.ustc.edu.cn}
\author[ahau,equal]{Tao Xu} 
\ead{taoxu@stu.ahau.edu.cn}
\author[ahau]{Xiaodi Li}
\ead{lixiaodi@stu.ahau.edu.cn}
\author[ahau]{Qian Gao}
\ead{gaoqian@stu.ahau.edu.cn}
\author[ustc]{Zhiwei Xiong}
\ead{zwxiong@ustc.edu.cn}
\author[ahu,corres]{Junfeng Xia}
\ead{jfxia@ahu.edu.cn}
\author[ahau,corres]{Zhenyu Yue}
\ead{zhenyuyue@ahau.edu.cn}

\affiliation[ustc]{organization={University of Science and Technology of China},
            city={Hefei},
            postcode={230026}, 
            state={Anhui},
            country={China}}
\affiliation[ahau]{organization={School of Information and Artificial Intelligence, Anhui Agricultural University},
            city={Hefei},
            postcode={230036}, 
            state={Anhui},
            country={China}}
\affiliation[ahu]{organization={Institutes of Physical Science and Information Technology, Anhui University},
            city={Hefei},
            postcode={230036}, 
            state={Anhui},
            country={China}}
\fntext[equal]{Equal contribution.}
\fntext[corres]{Corresponding authors.}

\begin{abstract}
Predicting the response of a cancer cell line to a therapeutic drug is pivotal for personalized medicine. Despite numerous deep learning methods that have been developed for drug response prediction, integrating diverse information about biological entities and predicting the directional response remain major challenges. Here, we propose a novel interpretable predictive model, DRExplainer, which leverages a directed graph convolutional network to enhance the prediction in a directed bipartite network framework. DRExplainer constructs a directed bipartite network integrating multi-omics profiles of cell lines, the chemical structure of drugs and known drug response to achieve directed prediction. Then, DRExplainer identifies the most relevant subgraph to each prediction in this directed bipartite network by learning a mask, facilitating critical medical decision-making. Additionally, we introduce a quantifiable method for model interpretability that leverages a ground truth benchmark dataset curated from biological features. In computational experiments, DRExplainer outperforms state-of-the-art predictive methods and another graph-based explanation method under the same experimental setting. Finally, the case studies further validate the interpretability and the effectiveness of DRExplainer in predictive novel drug response. Our code is available at: https://github.com/vshy-dream/DRExplainer.
\end{abstract}




\begin{keyword}
Drug response prediction \sep Directed graph convolutional network \sep Multi-omics \sep Quantifiable interpretability


\end{keyword}

\end{frontmatter}



\section{Introduction}
\label{sec1}


The success of drug development significantly influences advancements in modern medicine~\cite{manzari2021targeted}. In modern medical practice, accurately predicting an individual’s specific response while minimizing adverse reactions has become one of the key challenges in achieving personalized medicine and enhancing therapeutic efficacy \cite{weinshilboum2017pharmacogenomics}. Predicting the response of cancer cell lines to therapeutic drugs is a cornerstone of precision medicine, where treatments are tailored to individual patients based on their unique genetic profiles. This is particularly crucial in oncology, where inter-individual variability in drug responses can lead to ineffective treatments or severe adverse effects. Nevertheless, the field of drug response prediction encounters multifaceted challenges. The interplay of genetic diversity and the variation of environmental influences bring uncertainty to drug response prediction \cite{lu2014personalized}. Coalesced with these genetic traits are the dynamic layers of gene expressions, whose complexity shapes the variability of drug responses.

Parallel with these complexities, traditional prediction methodologies struggle with saving resources, often expensive and time-consuming. In recent years, notable progress in pharmacogenomics has yielded a wealth of genomic data, including gene mutations, copy number variations, gene expression profiles, and extensive datasets on drug sensitivity and resistance across different cancer cell lines \cite{barretina2012cancer,garnett2012systematic,basu2013interactive}. These datasets facilitate the development of computational models that can utilize multi-omics for drug response prediction. Within this condition, the integration of bioinformatics and machine learning opens up fresh research directions for drug response prediction. Previous studies have developed a range of models, including ridge regression \cite{liu2020improved}, support vector machine \cite{huang2018applications}, and transfer learning \cite{dhruba2018application}, to leverage available data for quantifying responses based on half-maximal inhibitory concentration (IC\textsubscript{50}) values. Lower IC\textsubscript{50} value indicates drug for sensitivity could confer a significant response to the specific targeted cell line \cite{carr2016defining}. Conversely, higher IC\textsubscript{50} value indicates a lack of significant drug response relationship. While these approaches have demonstrated encouraging performance, they often lack comprehensive incorporation of biological knowledge, constructing predictive models solely reliant on cell line gene expression data. Due to the wealth of available data recently, an increasing number of models are taking into cell line multi-omics along with drug chemical structure features \cite{peng2021predicting,sharifi2019moli,xu2024reusability}.

The field of graph-based learning has witnessed rapid-developed interest among researchers, leading to its rapid-developed application in the realm of bioinformatics \cite{li2022graph,shen2023systematic,shi2022gra,hasib2023depression}. Notably, these techniques have demonstrated remarkable utility in understanding complex biological systems. While graph-based learning methodologies have burgeoned across various domains \cite{xia2021graph,kipf2016semi,shishir2021novo}, the majority of approaches in drug response prediction have primarily operated within the confines of undirected bipartite networks. For example, Liu et al. utilized a hybrid graph convolutional network and multiple subnetworks for accurate prediction of cancer drug response \cite{liu2020deepcdr}, Liu et al. constructed a graph neural network with contrastive learning to enhance the generalization ability of drug response prediction \cite{liu2022graphcdr}, Zhu et al. proposed a framework consists of twin graph neural network and a similarity augmentation for drug response prediction \cite{zhu2022tgsa}. Typically, a thorough investigation of drug response demands a separate analysis of sensitivity and resistance. Thus, the necessity to advance algorithms for directed bipartite networks becomes apparent.

Furthermore, model interpretability has emerged as a critical concern. Despite significant advances, existing predictive models often lack the interpretability required for clinical adoption, limiting their utility for medical decision-making. In clinical practice, stakeholders, including physicians and patients, require not only accurate predictions but also a clear understanding of how and why a model arrives at a specific prediction. Many high-performing deep learning models, such as deep neural networks, are often regarded as ``black boxes", impeding their interpretability \cite{zhang2021survey}. Recently, several approaches have been proposed to explain the predictions based on graphs since graph neural networks have become increasingly popular \cite{zhang2020deep,jimenez2020drug}. This interpretability improves their adoption and trustworthiness in critical medical decision-making. Current efforts to enhance the interpretability of drug response prediction models predominantly rely on algorithms designed for homogeneous networks \cite{yuan2022explainability}. For example, researchers explored methodologies for identifying and ranking the most influential features or nodes within homogeneous networks. Huang et al. designed a nonlinear model to explain the node in the subgraph, computing the k most representative features \cite{huang2022graphlime}. Vu et al. proposed a probabilistic graphical model to identify the crucial nodes and the dependencies of explained features in the form of conditional probabilities \cite{vu2020pgm}. There are also explanation methods that identify exiting links in the network to explain the predicted links based on gradient or mutual information, such as GNNExplainer \cite{ying2019gnnexplainer} and ExplaiNE \cite{kang2019explaine}. Some studies delved into the identification of significant pathways and modules within homogeneous networks. However, there is a significant absence of interpretability methods tailored for directed bipartite networks. Additionally, the absence of standardized approaches that can evaluate the level of interpretability across various methodologies hampers the rigorous assessment of interpretability.

Evaluation of interpretability algorithms, a crucial task, aims to ensure that the predictions generated by the model are fully comprehensible to both the scientific community and clinical practitioners. However, the current evaluation of interpretability algorithms faces certain limitations and challenges. Existing evaluation methods primarily focus on qualitative assessment \cite{agarwal2023evaluating}, lacking a universal set of quantitative metrics to measure the quality and effectiveness of interpretability \cite{minh2022explainable}. In this regard, constructing a reliable ground truth is crucial. By establishing a truthful and credible ground truth, we can evaluate the accuracy and rationality of the interpretability algorithms concerning the model predictions. This construction process necessitates the integration of extensive domain knowledge and empirical evidence \cite{hao2022knowledge}. This integration enables an objective assessment of explainable algorithms’ strengths and weaknesses \cite{kuenzi2020predicting}. For example, taking into account the pathogenic mechanisms of cancer and cancer genes within the biomedical domain serves to promote their practical applications. In the realm of drug response prediction, considering interpretability algorithms that incorporate insights into prior knowledge and domain knowledge is particularly vital. Such consideration not only rationalizes the prediction outcomes but also enhances the applicability of interpretability algorithms in clinical practice.

Existing drug response prediction methods face critical challenges, including limited interpretability, dependence on undirected networks, and the lack of standardized evaluation approaches for interpretability. In this study, we presented DRExplainer, a novel approach that bridges these gaps by integrating a directed graph convolutional framework with a dedicated interpretability algorithm. Furthermore, we constructed a ground truth benchmark dataset to quantitatively evaluate the interpretability of the model, which sets it apart from existing approaches. Our main contributions are summarized as follows:
\begin{itemize}
\item We devised a novel approach that leverages directed graph convolutional network (DGCN) to address the challenges inherent in drug response prediction. Our approach harnessed the capabilities of DGCN, paving the way for more accurate and interpretable predictions in the challenging field of drug response.
\item We presented a paradigm by introducing a novel interpretability algorithm explicitly designed to tackle the complexities of directed bipartite networks in drug response prediction.
\item We constructed a ground truth benchmark dataset for each drug response pair by leveraging domain-specific knowledge, incorporating insights into the pathogenic mechanisms of cancer, relevant cancer genes, and the unique features of the drugs.
\item We introduced a quantitative approach for assessing explanations, offering a means to quantify the interpretability of our methodology. This quantitative framework enables a more profound understanding of the credibility and transparency of deep learning applications in the context of precision medicine.
\end{itemize}

\section{Materials and Methods}\label{sec2}

\subsection{Materials}\label{subsec1}
Our study leveraged a combination of diverse data sources, carefully curated to ensure the completeness and precision of the information used in our research. This study mainly involves two databases, Genomics of Drug Sensitivity in Cancer (GDSC) \cite{yang2012genomics} and Cancer Cell Line Encyclopedia (CCLE) \cite{barretina2012cancer}.  CCLE stands as a rich repository of multi-omics data from diverse cell lines, including genomics, transcriptomics, proteomics, and epigenomics. However, it's essential to note that the CCLE dataset is relatively limited in terms of drug response data. In contrast, GDSC primarily focuses on capturing drug response information, offering a valuable collection of drug sensitivity and resistance profiles across a diverse spectrum of cancer cell lines. So, we harnessed the multi-omics profiles, which encompass gene expression, gene mutation, and copy number variation, from the CCLE dataset. Simultaneously, we collected response scores of IC\textsubscript{50} values (natural log-transformed) from the GDSC dataset. In subsequent experiments, we used IC\textsubscript{50} values as a measure of sensitivity.

Following meticulous data collection, we obtained 986 cell lines, 358 drugs and a total of 295992 drug response data. During the sequencing and data collection of cell lines, it is common to encounter missing values. To ensure data quality and completeness, we focused on cell lines that possessed the three omics data types (i.e., gene expression, gene mutation and copy number variation). In order to thoroughly investigate the cancer-related mechanisms within drug response, we preprocessed the cell line omics data. Specifically, we retained the gene of cell line omics data listed in the COSMIC Cancer Gene Census \cite{wang2009pubchem}. This preprocessing resulted in feature vectors of dimensions 674, 689, and 480 for gene expression, gene mutation, and copy number variation, respectively. For drug data, we obtained the SMILES strings of related drugs from PubChem \cite{wang2009pubchem} and DrugBank \cite{knox2024drugbank} datasets. It is crucial to acknowledge that not all drugs are encompassed within these two databases. Table~\ref{tab1} presents a statistical overview of the collected dataset. After the collection of 157 drugs, we employed the ConvMolFeaturizer method in the DeepChem library \cite{duvenaud2015convolutional} to compile each drug’s SMILES string into a molecular graph. In this representation, nodes and edges correspond to chemical atoms and bonds, respectively. This systematic conversion enables a structured and graph-based representation of each drug’s molecular composition.

After rigorous preprocessing and data alignment, the final curated dataset encompasses 477 cancer cell lines and 157 drugs. As mentioned above, we utilized IC\textsubscript{50} values to measure the response between cell lines and drugs in this study. To categorize drug responses, we implemented two threshold values based on IC\textsubscript{50} values. Instances with IC\textsubscript{50} values \textless{} -3 were classified as sensitive responses, while those with IC\textsubscript{50} value \textgreater{} 3 were regarded as resistant. In a subsequent phase of our study, we compiled an expansive dataset featuring 17,071 drug responses, comprising 1,000 instances classified as sensitive and 16,071 as resistant. Specifically, we conducted the intricate drug response relationships within a directed graph framework, employing two adjacency matrices for representation. One adjacency matrix denoted sensitivity, where a value of 1 indicated a sensitive response, while the other matrix denoted resistance, with a value of 1 signifying a resistant response.

\begin{table}[!b]
\centering
\caption{A statistical summary of our dataset\label{tab1}}%
\begin{tabular*}{0.6\columnwidth}{@{\extracolsep\fill}llll@{\extracolsep\fill}}
\toprule
\multicolumn{2}{c}{Specification} & Collected  & Used\\
\midrule

\multirow{3}{*}{Cell line} & Gene expression & 644 &  \\
                            & Gene mutation & 798 & 477 \\
                            & Copy number variation & 605 &  \\ 
                            \addlinespace
\multirow{2}{*}{Response}   & Sensitive & 1000 & 1000 \\
                            & Resistant & 16071 & 16071 \\ \addlinespace
\multirow{2}{*}{Similarity} & Among Cell lines & NA & 41465 \\
                            & Among Drugs & NA & 3512 \\ \addlinespace
Drug                        &  & 358 & 157 \\ 
\bottomrule
\end{tabular*}
\end{table}

\begin{figure*}[!t]%
\centering
\includegraphics[width=\textwidth]{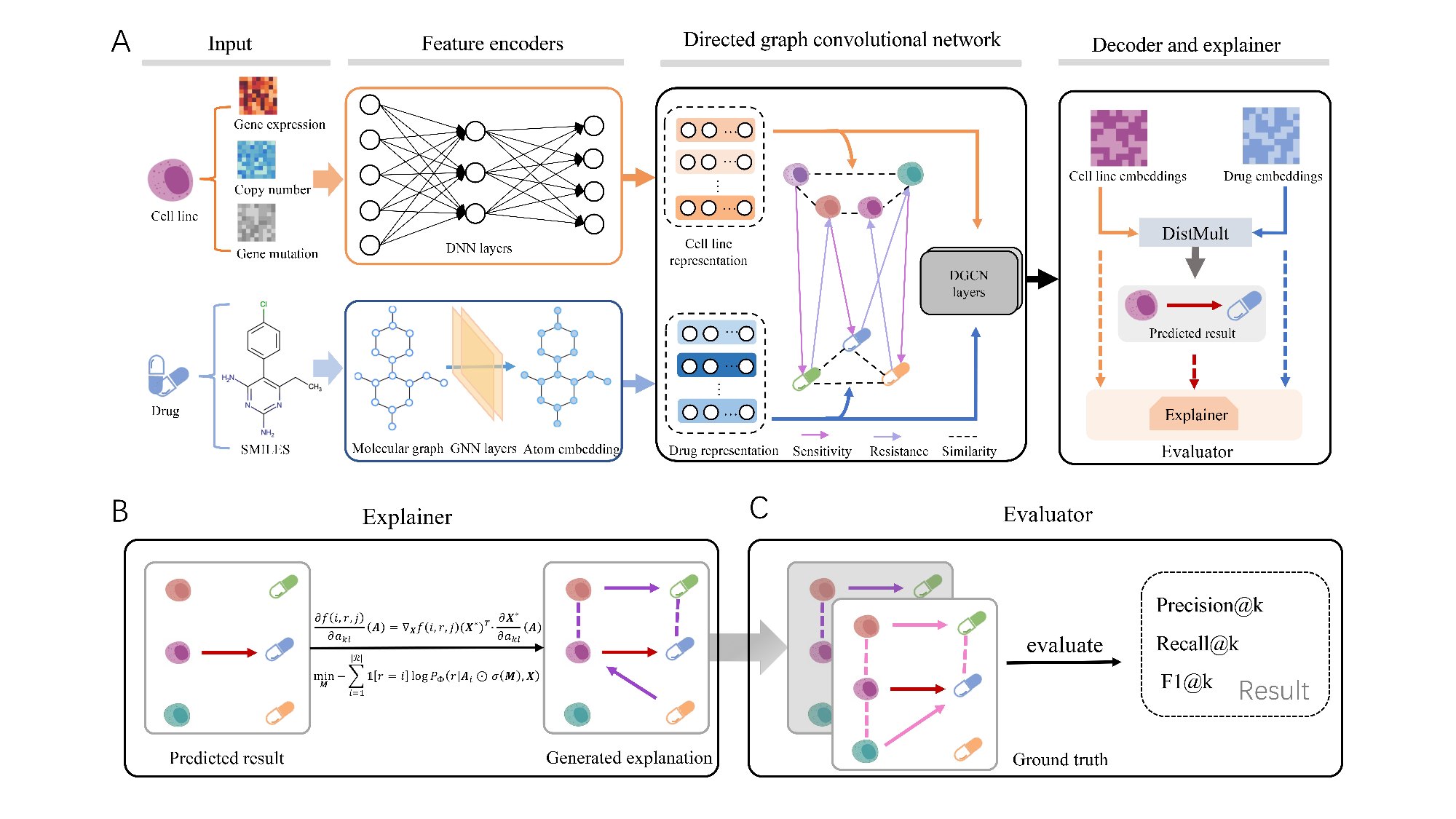}
\caption{Overview of DRExplainer framework. (A) DNNs and GNNs separately encode the input SMILES and multi-omics profiles. The encoded drug and cell line representations are fed into our directed graph convolutional network to learn further node embeddings. The updated embeddings are decoded by the DistMult decoder to predict novel responses. After prediction, the explainer and evaluator modules are employed to explain the prediction and evaluate the explanation, respectively. (B) The overview of explainer architecture. (C) The overview of evaluator architecture}\label{fig1}
\end{figure*}

\subsection{Predicting drug response}\label{subsec2}

As depicted in Figure~\ref{fig1}, our method, denoted as DRExplainer, is structured into three following modules: (A) the predictor, (B) the explainer, and (C) the evaluator. And module (A) predictor consisted of three main parts. Firstly, integrate the multi-omics profiles of cancer cell lines via deep neural network (DNN) layers and extract drug characteristics from the molecular graphs via graph neural network (GNN) layers. Subsequent to the integration, a directed bipartite network was constructed, synthesizing information from cell line similarity network, drug similarity network and cell line-drug directed network.

Exploiting the directed graph neural network, we updated the node representation within this network, aggregating both neighborhood information and multi-relational information. In the last period, we made predictions by DistMult decoder \cite{yang2014embedding} using the updated embeddings. Next, on the basis of the prediction, we devised an explainer to explain the prediction, aiming to understand the prediction, which was useful and vital in clinical research scenarios. Thereafter, a specific evaluator was adopted to assess whether the explanation was reasonable by comparing it with the pre-established ground truth benchmark dataset. The details of the explainer and evaluator will be described in Section ``Explaining the prediction" and Section ``Constructing ground truth and Evaluating". 

\subsubsection{Node representation module}\label{subsubsec1}
Inspired by the research of \cite{liu2020deepcdr}, we implemented omics-specific neural network layers to integrate multi-omics data, aiming to achieve distinctive representation for each cancer cell line. A full depiction of a cancer cell line’s multi-omics data including gene expression, gene mutation and copy number variant feature vectors denoted by ${\boldsymbol{c}_e}$, ${\boldsymbol{c}_m}$, $\boldsymbol{c}_n$. These vectors were subsequently transformed into an F-dimensional vectorial embedding through the following equation:
\begin{equation}
\boldsymbol{c}=g_c\left\{\left[g_e\left(\boldsymbol{c}_e\right)\left\|g_m\left(\boldsymbol{c}_m\right)\right\| g_n\left(\boldsymbol{c}_n\right)\right]\right\}
,\label{eq1}
\end{equation}
where $\boldsymbol{c} \in R^F$ denoted the representation of a cancer cell line, $\|$ standed for the vector concatenation operator. The term $\left\{g_c, g_e, g_m, g_n\right\}$ represents various neural network layers designed for feature transformation. Considering a set of cancer cell lines $C=\left\{c_i\right\}_{i=1}^{N_C}$, where $N_C$ indicated the total number of cell lines, all cancer cell line representations were ultimately obtained as $\boldsymbol{C} \in \mathbb{R}^{N_C \times F}$ to be used in subsequent modeling.

As described above, drugs were treated as molecular graphs, with nodes representing atoms and edges denoting the chemical bonds between them. The molecular graph of a specific drug was represented as $G_d=\left(\boldsymbol{X}_{\boldsymbol{d}}, \boldsymbol{A}_{\boldsymbol{d}}\right)$, where $\boldsymbol{X}_{\boldsymbol{d}} \in \mathbb{R}^{N_d \times F_d}$ was the node attribute matrix representing atomic vectors $\left(F_d=75\right)$, and $A_d \in \mathbb{R}^{N_d \times N_d}$ was an adjacency matrix mapping chemical bonds between atoms. Here, $N_d$ signified the count of atoms within the molecular graph of drug $d$. In this study, we leveraged a graph neural network (GNN) encoder, symbolized as $\phi$, to capture the latent representation of atom nodes. Given a set of drugs $D=\left\{d_i\right\}_{i=1}^{N_D}$, where $N_D$ denoted the total number of drugs, we engaged the GNN encoder $\phi$ to extract a comprehensive representation $\boldsymbol{D} \in \mathbb{R}^{N_D \times F}$ of all drugs for the following modeling.

\subsubsection{Cell line and drug similarity}\label{subsubsec2}
Building on the premise that cell lines with similar properties exhibit comparable drug sensitivities \cite{guan2019anticancer,zhang2015predicting}, we incorporated cell line and drug similarity as key attributes for drug sensitivity inference. Hence, we computed the similarity among cell lines and drugs, relying on the representations obtained in the section ``Node representation module". These similarities were quantified using cosine similarity, which performed pairwise global alignment among cell lines or drugs. The cell line-cell line similarity and the drug-drug similarity network were denoted by the matrices $\boldsymbol{S}^c=\left[S_{i j}^c\right] \in \mathbb{R}^{N_c \times N_c}$ and $\boldsymbol{S}^d=\left[S_{i j}^d\right] \in \mathbb{R}^{N_d \times N_d}$, respectively. Here, $S_{i j}^c$ and $S_{i j}^d$ quantified the pairwise similarities between cell lines $c_i$ and $c_j$, and between drugs $d_i$ and $d_i$. The total number of cell lines and drugs were denoted by $N_r$ and $N_f$. In this investigation, we employed cosine similarity to measure the similarity among cell lines or drugs, with the similarity score represented as $\cos s_{i j}$, which evaluates the degree of similarity between cell lines $c_i$ and $c_i$ or drugs $d_i$ and $d_j$. The cosine similarity was calculated as follows:
\begin{equation}
\cos s_{i j}=\cos \left(\boldsymbol{x}_{\boldsymbol{i}}, \boldsymbol{x}_{\boldsymbol{j}}\right)=\frac{\boldsymbol{x}_{\boldsymbol{i}} \cdot \boldsymbol{x}_{\boldsymbol{j}}}{\left|\boldsymbol{x}_{\boldsymbol{i}}\right| \cdot\left|\boldsymbol{x}_{\boldsymbol{j}}\right|}
,\label{eq2}
\end{equation}
where $\boldsymbol{x}_{\boldsymbol{i}}$ and $\boldsymbol{x}_{\boldsymbol{i}}$ represented the feature vectors of entity $i$ and $j$, respectively. To mitigate noise from weak similarities, we introduced a threshold, denoted by $\varphi$, to constrain the distribution of the similarity data. Specifically, we retained only similarity values where $\varphi>0.9$ for cell lines and $\varphi>0.88$ for drugs, thereby focusing on the most significant relationships. Hence, the similarity matrix can be obtained as follows:
\begin{equation}
S_{i j}=\left\{\begin{array}{l}
1, \text { if } \cos s_{i j} \geq \varphi \\
0, \text { otherwise }
\end{array}\right.
,\label{eq3}
\end{equation}

\subsubsection{Directed graph convolutional neural netwrok}\label{subsubsec3}
The directed bipartite network contained diverse types of relationships between distinct nodes. To more effectively integrate neighborhood information and capture the network structure, drawing inspiration from Schlichtkrull et al. \cite{schlichtkrull2018modeling}, we proposed the directed graph convolutional neural network for the enhancement of node representations. Within this framework, we leveraged the directed edges for the propagation and aggregation of information, respecting the directional flow inherent in biological and chemical entities. Our model iteratively updated node representations within the directed multi-graphs as $G=(\mathcal{V}, \mathcal{E}, \mathcal{R})$, where $\mathcal{V}$ represented the set of vertices, $\mathcal{E}$ consisted the set of edges and $\mathcal{R}$ signified the set of relations. As mentioned in the previous section, the directed heterogeneous network encompassed cell line-cell line similarity, drug-drug similarity, cell line-drug sensitive interaction, and drug-cell line resistant interaction. Consequently, the directed multi-graphs were characterized by four distinct types of edges. Then, $\mathcal{G}$ can be further expressed by a set of adjacency matrices $\boldsymbol{\mathcal{A}}=\left\{\boldsymbol{A}_i\right\}_{i=1}^{|\mathcal{R}|}$ and node attributes $\boldsymbol{X} \in$ $\mathbb{R}^{|\nu| \times F}$, where $\boldsymbol{A}_1$ and $\boldsymbol{A}_2$ corresponded to the sensitivity and resistance matrices discussed in the Section ``Materials", $\boldsymbol{A}_3$ and $\boldsymbol{A}_4$ equate to $\boldsymbol{S}^c$ and $\boldsymbol{S}^d$ mentioned in the Section ``Cell line and drug similarity", respectively, and $\boldsymbol{X}=$ $\left[\begin{array}{l}\boldsymbol{C} \\ \boldsymbol{D}\end{array}\right] \in \mathbb{R}^{\left(N_C+N_D\right) \times F}$ (i.e., the previously learned representations of cell line, $\boldsymbol{C}$, and drugs $\boldsymbol{D}$, were combined as $\boldsymbol{X}$).

In this paper, we utilized triples (subject, relation, object) to represent the drug response facts. Employing the directed graph convolutional neural network as an encoder, we updated the feature representation of the subject and object nodes. Motivated by the architecture \cite{schlichtkrull2018modeling}, we formulated the subsequent straightforward layer-wise propagation model to facilitate the computation of the forward-pass update for a node denoted by $v_i$ within the directed multi-graphs:
\begin{equation}
\boldsymbol{X}_i^{(l+1)}=\sigma\left(\sum_{r \in \mathcal{R}} \sum_{j \in \mathcal{N}_i^r} \frac{1}{c_{i, r}} W_r^{(l)} \boldsymbol{X}_j^{(l)}+W_0^{(l)} \boldsymbol{X}_i^{(l)}\right)
,\label{eq4}
\end{equation}
where $\boldsymbol{X}_i^{(l)} \in \mathbb{R}^{F^{(l)}}$ represented the feature matrix of node $v_i$ at the $l$-th layer with $F^{(l)}$ being the dimensionality of the layer, $\mathcal{N}_i^r$ designated the set of neighbor indices for node $v_i$ that were associated through relation $r \in \mathcal{R}$,$ W_0^{(l)}$ and $W_r^{(l)}$ were trainable weight matrices at the $l$-th layer, and $\sigma$ refered to the sigmoid activation function. According to formula (4), the feature representation of each node within the graph can be updated in parallel. Thus, we implemented a two-layer DGCN model to update the node representation. Subsequently, we introduced a decoder (scoring function) $f(s, r, o)$, which assigns scores to assess to evaluate how likely the edges $(s, r, o)$ being factual in the following section.

\subsubsection{DistMult decoder}\label{subsubsec4}
Taking the final cell line and drug embedding learned by the encoder as input, we introduced DistMult factorization \cite{yang2014embedding} as a decoder to work out the link prediction. By processing triples of the form (subject, relation, object), DGCN captured the inherent structure and relational context of the graph, thus providing a robust foundation for the subsequent decoding process.

The DistMult model operated on the assumption that the probability of a factual relationship between entities can be estimated through a simple bilinear form. In this model, each relation type $r$ was associated with a trainable diagonal matrix $\boldsymbol{N}_r \in \mathbb{R}^{d \times d}$, effectively reducing the number of relation parameters. For a given triple $(s, r, o)$, the DistMult decoder calculated a score representing the probability of this triple being factual. The score is computed as follows:
\begin{equation}
f(s, r, o)=\sigma\left(\boldsymbol{X}_s^T \boldsymbol{N}_r \boldsymbol{X}_o\right)
,\label{eq5}
\end{equation}

Regarding the model's overall architecture, DRExplainer can be characterized as $\Phi$ learning a conditional distribution $P_{\Phi}(r\mid\mathcal{A}, \boldsymbol{X})$. As in previous work on factorization \cite{trouillon2016complex}, we trained the model with negative samples. The positive samples in the training set were often randomly corrupted to generate negative samples. This approach will result in the generated negative sample being consistent or paradoxical with the positive sample. Inspired by these investigations \cite{rossi2021knowledge, hasib2020survey, hasib2021hsdlm}, we generated negative triples by treating the unutilized data in the data space as negative samples. To enhance model performance, we employed cross-entropy loss optimization, aiming to encourage the model to assign higher scores to observable triples compared to negative ones. The loss function was mathematically expressed as:
\begin{equation}
\mathcal{L}=-\frac{1}{|\mathcal{T}|} \sum_{(s, r, o, y) \in \mathcal{T}}\left(\begin{array}{c}
y \log l(f(s, r, o))+ \\
(1-y) \log (1-l(f(s, r, o)))
\end{array}\right)
,\label{eq6}
\end{equation}
where $\mathcal{T}$ represented the training set comprising both real and artificially generated negative triples, and $l$ denoted the logistic sigmoid function. Additionally, $y$ served as a binary indicator where $y=1$ corresponded to positive triples and $y=0$ to negative ones.

\subsection{Explaining the prediction}\label{subsec3}
Currently, investigating the interpretability of models using graph convolutional architectures has received increasing attention in graph representation learning \cite{yuan2022explainability}. Drawing inspiration from Ying et al.~\cite{ying2019gnnexplainer}, we developed DRExplainer to generate explanations for drug response prediction in the bipartite network. DRExplainer can explain prediction by seamlessly integrating both graph structure and node feature information. Initially, we meticulously generated ground truths data for individual drug response pairs, leveraging biological prior knowledge. The methodology for constructing ground truths was detailed in the Section ``Constructing ground truths and Evaluating". Our analysis prioritized explanations based on linked pairs, as opposed to unlinked pairs. We posited that such positive explanations yield greater insight, particularly in the context of biological networks, compared to negative explanations.

DRExplainer concentrated on the identification of the most relevant subgraphs within the directed bipartite network by learning a mask over the input adjacency matrix. The method involved minimizing the cross-entropy between the predicted label using the original input adjacency matrix and the predicted label derived from the adjacency matrix after applying the mask. This approach enhanced the comprehension of influential substructures within the network. The objective function minimized by DRExplainer was as follows:
\begin{equation}
\min _{\boldsymbol{M}}-\sum_{i=1}^{|\mathcal{R}|}\mathds{1}[r=i] \log P_{\boldsymbol{\Phi}}\left(r \mid \boldsymbol{A}_i \odot \sigma(\boldsymbol{M}), \boldsymbol{X}\right)
,\label{eq8}
\end{equation}
where $\boldsymbol{M}$ was the mask to be learned, $\mathds{1}$ denoted the indicator function and $\odot$ symbolized the element-wise multiplication applied to $\sigma(\boldsymbol{M})$ and the adjacency matrix $\boldsymbol{A}_i$. Subsequently, we applied a threshold to the product to filter out the low-value elements. This process complicated the extraction of the explanatory subgraph $G_s$. 

\subsection{Constructing ground truths and evaluating}\label{subsec4}

\begin{figure}[!b]%
\centering
\includegraphics[width=230pt]{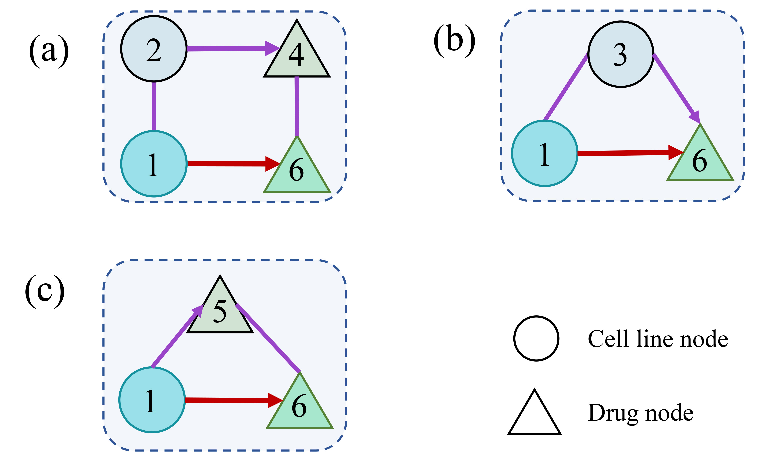}
\caption{Three ground truth construction methodologies for drug response prediction.}\label{fig2}
\end{figure}

While increasing research focuses on interpretability evaluation \cite{zhou2021evaluating,nauta2023anecdotal}, an evaluation framework for explaining bipartite network interactions remained underdeveloped. We were advancing a quantifiable paradigm for interpretability and devising a corresponding evaluation algorithm. To establish a robust framework for evaluation, our efforts concentrated on creating a definitive ground truth dataset.

The construction of a reliable ground truth for drug response relationships was a fundamental step in developing predictive models for personalized medicine. Our proposed method was built on the premise that cell lines with similar characteristics exhibit comparable drug sensitivities \cite{guan2019anticancer,sheng2015optimal}. We leveraged the interactions between entity similarity within the directed bipartite graph network to model the evaluation framework, as illustrated in Figure~\ref{fig2}. Figure~\ref{fig2} depicts three robust strategies to establish ground truths for the interaction between a specific drug (node 6) and a target cell line (node 1). Line segments in Figure~\ref{fig2} represented entity similarities as defined in the Section ``Cell line and drug similarity", while directed arrows indicated the directional drug response. For clarity, our methodology differentiated cell lines and drugs using circular and triangular nodes, respectively. The details of our methodologies are described below:

Method (a) generated ground truths by considering similarities across both cell lines and drugs. Nodes 2 and 4 were similar to nodes 1 and 6, respectively, and had the same response profiles. Consequently, the directed edge from nodes 2 to 4, along with the associated similarity edges, constituted the ground truths for the observed drug response interaction between nodes 1 and 6. Method (b) and method (c) employed a straightforward approach to constructing ground truths, in which a cell line or drug was similar to the related nodes and exhibited a consistent association with the observed drug response. 

Following the methodology, we established a benchmark dataset of drug response with ground truths. Building on the ground truth benchmark dataset, we introduced a quantifiable evaluation framework that utilizes recision@k, recall@k and f1@k to rigorously assess the quality of model explanations in drug response predictions. This framework evaluated the ranked lists of explanations produced by the explanatory algorithm. Specifically, precision@k measured the proportion of relevant explanations found in the top-k positions of the ranked predicted results, offering insights into the accuracy of the interpretable model. Recall@k quantified the model's capacity to capture all relevant explanations, determined by the number of ground truths identified within the top-k predictions. Meanwhile, f1@k provided a balanced measure of precision and recall, offering an integrated metric that assessed the model’s explanatory performance. The mentioned metrics were respectively calculated as follows:
\begin{equation}
\text {precision@} \mathrm{k}=\frac{R G}{k}
,\label{eq9}
\end{equation}
\begin{equation}
\text {recall@} \mathrm{k}=\frac{R G}{T G}
,\label{eq10}
\end{equation}
\begin{equation}
\mathrm{f} 1 @ \mathrm{k}=\frac{2 * \text {precision@} k * \text {recall@k}}{\text {precision@}k+ \text {recall@} k}
,\label{eq11}
\end{equation}
where RG denoted the count of the top-k generated explanations among relevant ground truths, while TG depicted the total number of generated explanations within the ground truths benchmark dataset. These metrics collectively provided a comprehensive assessment of the model’s explanatory capabilities, with higher values indicating better reliability and interpretative performance.

\begin{table}[b] 
\centering
\caption{Summary of corresponding drug responses' information in the test dataset of four tasks.\label{tab2}}
\begin{tabular*}{0.6\columnwidth}{@{\extracolsep\fill}llll@{\extracolsep\fill}}
\toprule
Experimental setting & Drug  & Cell line & Response\\
\midrule
TaskA    & Known   & Known  & Novel  \\
TaskB    & Known   & Novel  & Novel  \\
TaskC    & Novel   & Known  & Novel  \\
TaskD    & Novel   & Novel  & Novel  \\
\bottomrule
\end{tabular*}
\end{table}

\begin{figure}[t]%
\centering
\includegraphics[width=480pt]{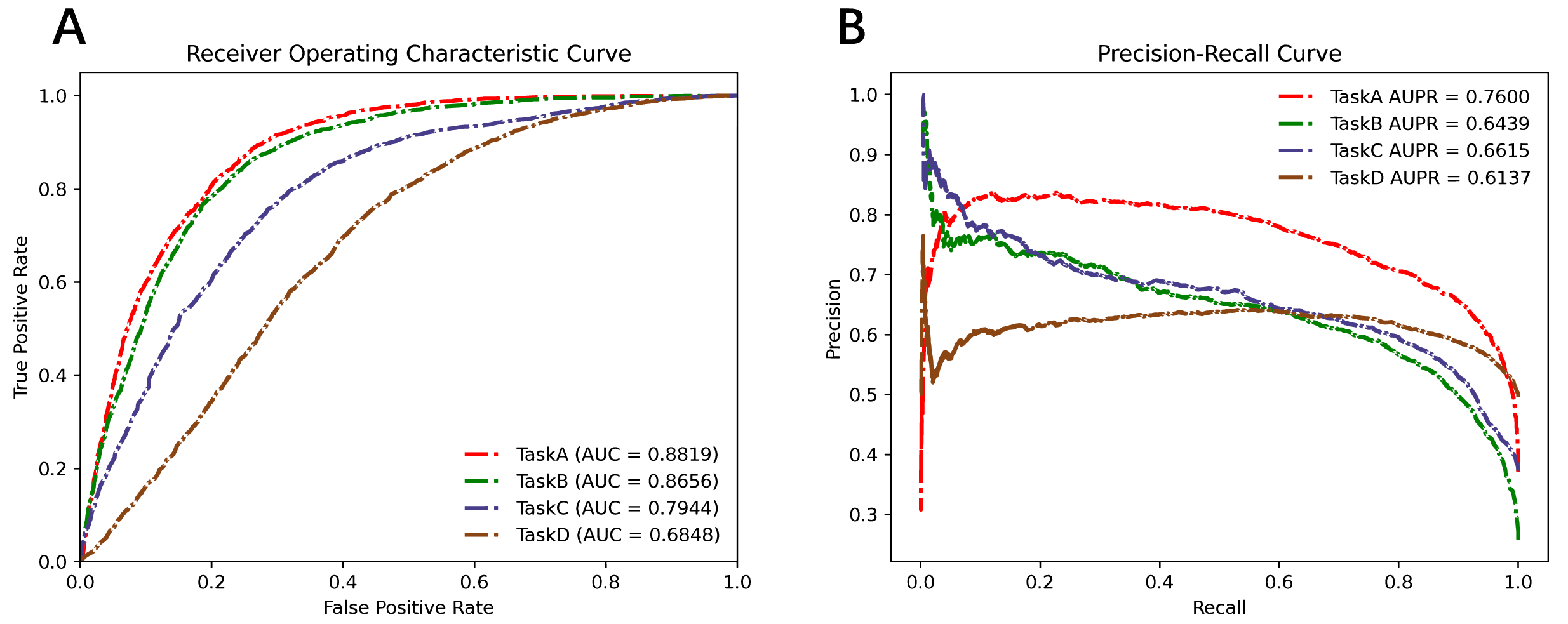}
\caption{The AUC and AUPR scores of DRExplainer in inductive capability learning.}\label{fig3}
\end{figure}


\begin{figure*}[t]%
\centering
\includegraphics[width=480pt]{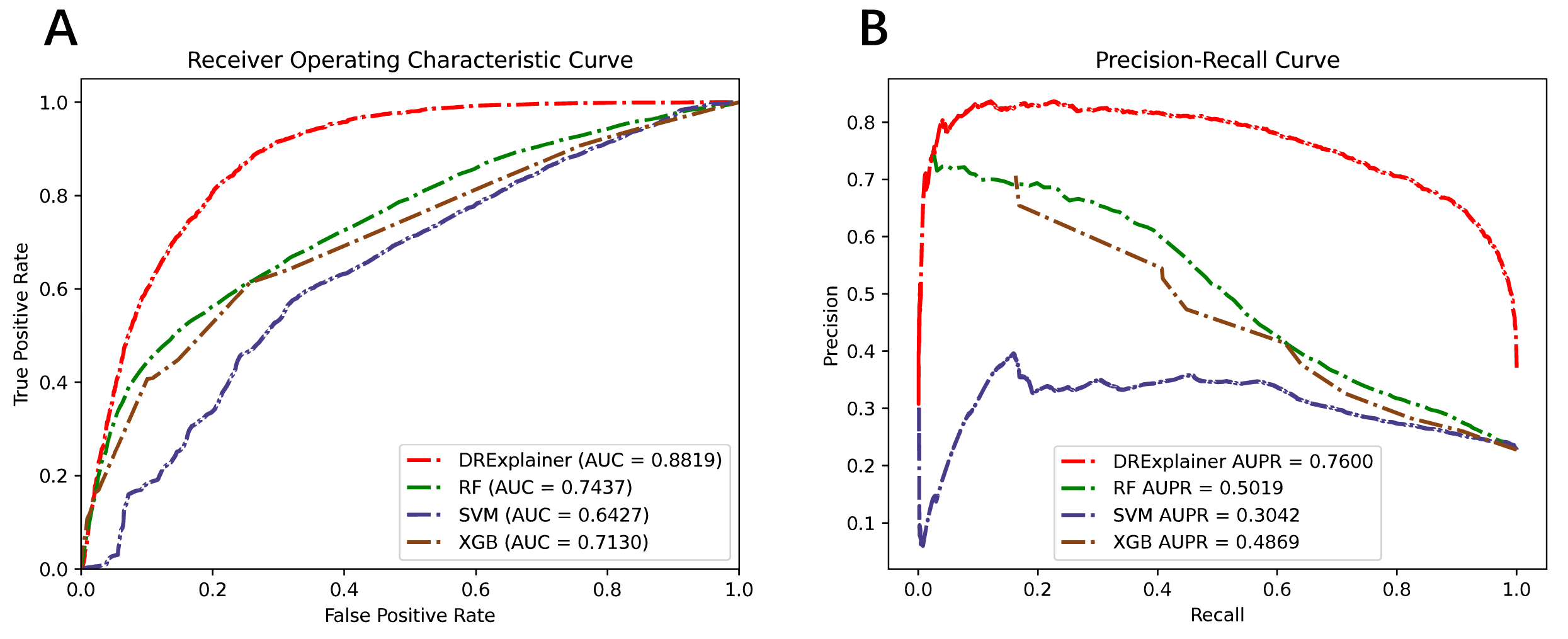}
\caption{The receiver operating characteristic and precision-recall curve of DRExplainer and baselines on the independent test.}\label{fig4}
\end{figure*}

\begin{figure*}[t]%
\centering
\includegraphics[width=480pt]{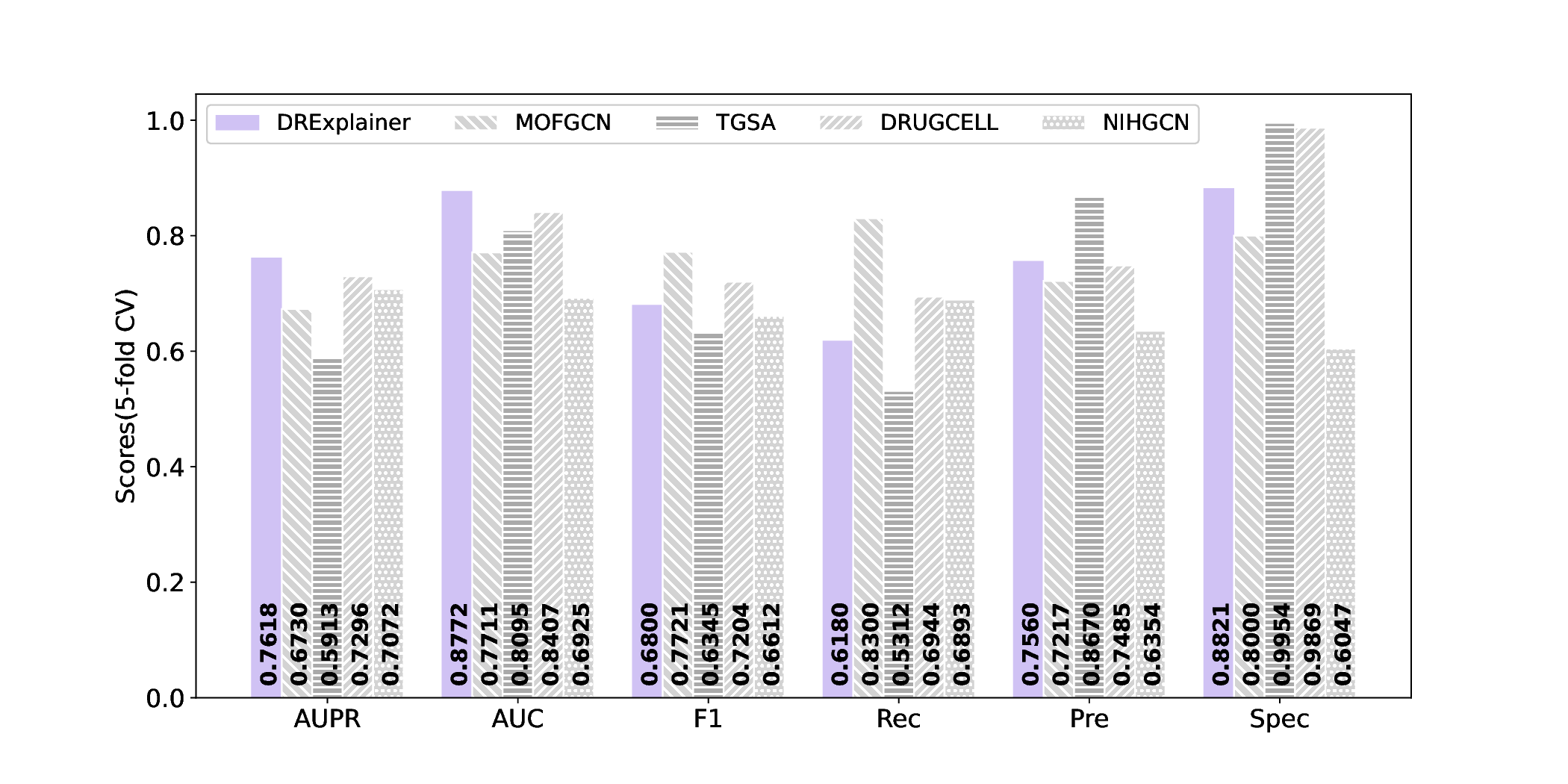}
\caption{The performance of DRExplainer and baselines on the cross-validation.}\label{fig5}
\end{figure*}

\section{Results}\label{sec3}
\subsection{Experimental design}\label{subsec5}
Our study advanced the field of drug response prediction by introducing a model capable of distinguishing the directionality of responses, including sensitivity, resistance and non-significant interaction. To underscore the robustness and generalizability of our model, we meticulously designed a data separation strategy, as summarized in Table~\ref{tab2}. The table outlined four distinct tasks, each with a unique distribution of drugs and cell lines within the test dataset. TaskA was the conventional experimental setting, predicting the novel responses with a test dataset composed of known drugs and cell lines. TaskB and TaskC evaluated the model’s capacity with novel cell lines and drugs, respectively. TaskD represented the most stringent test, where both drugs and cell lines in the test dataset are new to the model. 

Each task was rigorously tested using a 5-fold cross-validation to ensure the robustness and reliability of our findings. And Schlichtkrull et al. \cite{schlichtkrull2018modeling} have proved that incorporating inverse relation pairs is an effective solution for improving the model performance. Thus, we augmented our dataset by extending each triplet pair with corresponding inverse edges after data separation. This procedure ensured the absolute segregation of the training and test datasets, maintaining the fairness of our model evaluation.

Figure~\ref{fig3} delineated a comparative assessment of our model’s inductive capability across four distinct experimental tasks, which were designed to rigorously test the model’s predictive accuracy in various scenarios. Each task’s performance was illustrated via the Receiver Operating Characteristic (ROC) curve Precision-Recall Curve, providing insight into the model’s performance. TaskA adopted a traditional experimental setting, utilizing a test dataset composed of known drugs and cell lines to predict novel responses. And TaskA achieved the highest Area Under the Curve (AUC), indicating robust predictive ability within known entities. This setup served as a benchmark for the model’s performance under familiar conditions. In the analysis of inductive capability concerning cell lines and drugs, depicted in Task B and Task C, the performance of DRExplainer for cell lines exhibited a modest decline compared to drugs. This discrepancy may stem from the truth that cell lines share similar genetic information while chemical structures of drugs can be diversified. TaskD, while presenting the most challenging conditions, still showcased considerable predictive performance, albeit lower than the other tasks. Together, the results across all tasks underscored the model’s robust inductive capability, highlighting its potential utility in real-world clinical settings.

\subsection{Predicting method comparison}\label{subsec6}
To verify the effectiveness of our model DRExplainer, we conducted comparative analyses against several established machine learning algorithms, including support vector machine (SVM), random forest (RF), and XGBoost. In order to illustrate the superiority of our model proposed in this study, we also benchmarked against three state-of-the-art (SOTA) algorithms: MOFGCN \cite{peng2021predicting}, NIHGCN \cite{peng2022predicting}, DrugCell \cite{kuenzi2020predicting}, and TGSA \cite{zhu2022tgsa}, and the latter two of which were extensively discussed in the literature \cite{shen2023systematic}. Notably, MOFGCN, TGSA and NIHGCN are graph-based learning approaches widely applied in bioinformatics. Each algorithm underwent a five-fold cross-validation experiment on our dataset for a comprehensive performance assessment.

Figure~\ref{fig4} presented the comparative performance of various machine learning approaches, evaluated using the Area Under the Precision-Recall Curve (AUPR) and the Area Under the Receiver Operating Characteristic Curve (AUC). To ensure a fair and consistent comparison, these experiments were conducted using the feature derived from the Section "Node representation module". These features include multi-omics and molecular-level data, which are widely utilized in drug response prediction tasks. To elucidate the findings in the figure, we reported the results of the best performing fold. The results demonstrated that our model consistently surpassed the baselines algorithms, achieving the highest scores for both AUPR and AUC metrics. Further performance analysis revealed that conventional machine learning algorithms frequently encountered difficulties when processing a large amount of data and complex associations, as previously noted by Wang et al. \cite{wang2021comparative}, which contributed to their suboptimal performance relative to our model. More detailed experiments about machine learning algorithms are provided in the Supplementary material. To rigorously assess the efficacy of our model, we compared our method with the following SOTA methods.
\begin{itemize}
\item MOFGCN employsed multi-scale network integration to model the complex interactions between drugs and cell lines, leveraging the chemical properties of drugs and genetic attributes of cell lines for informed analysis.
\item DrugCell utilized drug response data from tumor cell lines in conjunction with genotype information to predict drug efficacy, highlighting potential synergistic effects or resistance mechanisms specific to cell lines.
\item TGSA capitalized on the complexity within protein-protein interaction networks to construct high-quality cell line representations, which were crucial for accurate drug response predictions based on genetic profiles.
\item NIHGCN leveraged an innovative heterogeneous graph convolutional network architecture, combining parallel graph convolution and neighborhood interaction layers to achieve precise predictions of anticancer drug responses.
\end{itemize}

Figure~\ref{fig5} delineated the performance across different metrics, providing a comprehensive evaluation of our model’s capabilities. The AUC served as a global measure of model performance, independent of any specific classification threshold, meanwhile, the AUPR provided a more detailed assessment, particularly beneficial for imbalanced datasets, by focusing on the precision-recall balance across varying thresholds. Together, these metrics were essential for a comprehensive evaluation of classification models, with higher values signifying superior class discrimination and predictive accuracy. In terms of both AUC and AUPR, our model exhibited outstanding performance, outperforming other SOTA models. Specifically, it exceeded the AUC scores of MOFGCN, TGSA, DrugCell, and NIHGCN by 13.76\%, 8.36\%, 4.34\%, and 26.67\%, respectively, and achieved similar margins of improvement in AUPR scores. These results highlighted the effectiveness and robustness of our model in drug response tasks.

In addition, our model showcased remarkable specificity, an essential trait in analyses where negative samples predominated over positive ones. Nonetheless, it was important to acknowledge that, despite its overall superiority, our model did not attain the highest F1 score when compared to MOFGCN and DrugCell. This divergence was ascribed to MOFGCN and DrugCell’s higher recall, which cumulatively contributed to their augmented F1 score. And our model’s ability to tackle complex biological problems, as mentioned above, was noteworthy. Moreover, we also conducted the inductive learning experiments of the SOTA methods and the results were provided in the Supplementary material. DRExplainer extended beyond the common prediction of sensitivity to accurately identify resistance and non-significant relationships, demonstrating its contribution to the field of drug response research.

\begin{figure}[!b]%
\centering
\includegraphics[width=350pt]{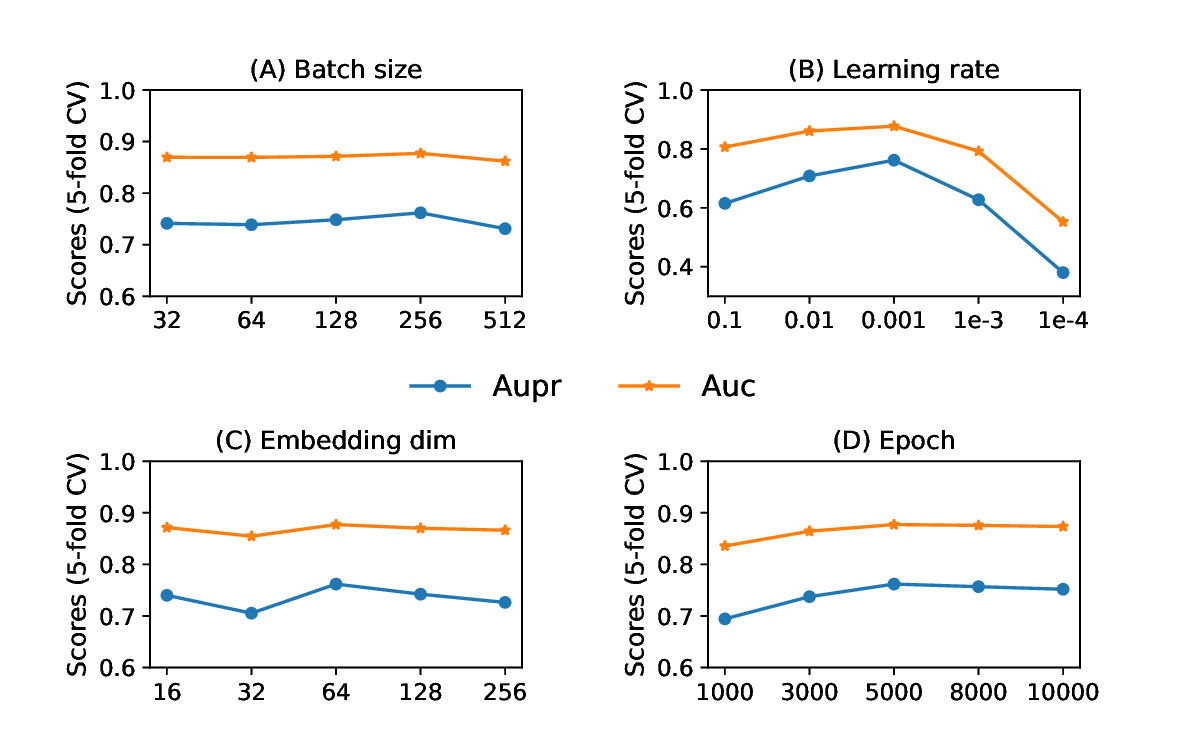}
\caption{Parameter sensitivity of DRExplainer w.r.t. (A) the batch size, (B) the learning rate, (C) the hidden embedding dimension, (D) the training epoch.}\label{fig6}
\end{figure}

\subsection{Parameter sensitivity analysis}\label{subsec7}
In this section, we conducted a thorough parameter sensitivity analysis to further study the influence of parameters on our model performance. Through grid search methodology, DRExplainer achieved the best results with batch size of 256, learning rate of 0.001, embedding dimension of 64 and the training epoch of 5000. Therefore, we used these optimal hyperparameters as the default configuration for subsequent sensitivity analysis of individual hyperparameter. This experiment was rigorously validated by a five-fold cross-validation framework, leveraging AUC and AUPR as the principal performance metrics. The comprehensive sensitivity analysis of our model was illustrated in Figure~\ref{fig6}.

Specifically, subgraphs (A), (C), and (D) of Figure~\ref{fig6} showed marginal variations in AUPR and AUC in terms of batch size, embedding dimension and epoch. This suggested that the model’s generalization ability was not substantially dependent on these parameters. Such stability was indicative of our model’s ability to maintain good performance, even varying in batch sizes, feature representation complexity, and training epochs. 

Conversely, our model exhibited sensitivity to learning rate as depicted in subgraph (B) of Figure~\ref{fig6}. A sharp decline in performance at both high and low learning rates signifies that the learning rate was a critical parameter that requires careful tuning. This sensitivity to the learning rate was consistent with its role in regulating the magnitude of weight updates throughout the optimization process, thereby influencing the model’s convergence and training stability.

In summary, the analysis demonstrated the robustness of the model against several parameters, emphasizing its reliability in various configurations. However, the sensitivity to learning rate accentuated the significance of this parameter in achieving peak model performance, and therefore, it must be tuned with precision.

\subsection{Evaluating the explanation}\label{subsec8}
Interpretability is crucial in predictive modeling, especially in predicting drug responses, where comprehension of the underlying mechanism is as vital as the prediction itself. To further assess the interpretability of our DRExplainer, we conducted comparative experiments with another graph-based method, ExplaiNE. These evaluations were conducted on our curated ground truth benchmark dataset, providing a rigorous basis for testing the explanatory power of each model. 

ExplaiNE was employed to evaluate the impact of small alterations in the adjacency matrix on the scoring function, by computing the gradients of the model's scoring function concerning the adjacency matrix. The gradient computation allowed for the identification of critical edges in the network that significantly influenced the model's prediction. Specifically, consider a scenario involving pairs of nodes $(i, r, j)$ for potential prediction and pairs $(k, r, l)$ for candidate explanation. Within this framework, a specific score was allocated to the node pair $(k, r, l)$ as follows:
\begin{equation}
\frac{\partial f(i, r, j)}{\partial a_{k l}}(\boldsymbol{A})=\nabla_{\boldsymbol{X}} f(i, r, j)\left(\boldsymbol{X}^*\right)^T \cdot \frac{\partial \boldsymbol{X}^*}{\partial a_{k l}}(\boldsymbol{A})
,\label{eq7}
\end{equation}
where $\boldsymbol{X}^*$ was the optimal embedding for adjacency matrix $\boldsymbol{A}$, and $a_{k l}$ denoted an element of the adjacency matrix $\boldsymbol{A}$. 
Kang et al. \cite{kang2019explaine} elucidated that the most plausible explanation for a predicted link $(i, r, i)$ involving a node $i$ is often a link $(k, r, l)$ that connected to node $i$ (i.e., where $l=i$ or $k=i$ ). This was attributed to the fact that links neighboring node $i$ had a direct influence on the link probability $f(i, r, j)$ through their impact on the node embedding $\boldsymbol{X}_i^*$. Consequently, our analysis focused on the immediate, first-order connections adjacent to the two nodes constituting the interpreted edge. To ensure an equitable comparison, we only consider the first-order neighbors of the nodes of the triple in the subgraph for DRExplainer. 

According to Table~\ref{tab3}, DRExplainer significantly outperforms ExplaiNE across all metrics, including recall@k, precision@k and f1@k. The superior performance demonstrates the ability of DRExplainer to capture relevant graph information that contributes to model predictions. The efficacy of DRExplainer can be attributed to its more effective mechanism for identifying and interpreting the complex interactions with the graph data, thereby providing clearer insights into the decision-making processes of graph neural networks. In clinical practice, where the stakes of predictive accuracy and understanding are high, the employment of DRExplainer could be particularly advantageous in elucidating drug response mechanisms in a more comprehensive and precise manner.

\begin{table}[!t]
\centering
\caption{The performances of DRExplainer and ExplaiNE on the curated ground truth benchmark dataset.\label{tab3}}
\begin{tabular*}{0.6\columnwidth}
{@{\extracolsep\fill}llll@{\extracolsep\fill}}
\toprule
Method & Precision@k & Recall@k & F1@k \\
\midrule
DRExplainer & \textbf{0.4199} & \textbf{0.3740} & \textbf{0.4208} \\
ExplaiNE & 0.3979 & 0.2934 & 0.3335 \\
\bottomrule
\end{tabular*}
\end{table}

\subsection{Case study}\label{subsec9}
In this section, we presented two case studies to verify the effectiveness and interoperability of DRExplainer in drug response.

In Case Study 1, We trained DRExplainer with the known cell line-drug responses in the training set, and then assessed its predictions in the test set, which were prioritized by scores. The top 10 sensitive and resistant responses predicted by DRExplainer are illustrated in Table~\ref{tab4}.
As shown in Table~\ref{tab4}, most results can be verified based on the existing literature in PubMed. Here, we took two typical responses for analysis. Notably, OCIAML2 cell line is derived from acute myeloid leukemia and 5-Fluorouracil (5-FU) is a drug that inhibits thymidylate synthase and blocks DNA synthesis, leading to cell growth arrest and death~\cite{wigmore2010effects}. We predicted that OCIAML2 cell line is sensitive to 5-Fluorouracil drug (5-FU), aligning with findings by Falzacappa et al. that showed significant sensitivity (EC\textsubscript{50}=3$\mu$umol/L)~\cite{falzacappa2015combination}. Additionally, for human tumor cell lines OCILY19, which was predicted to have a resistant response with the drug Etoposide. This finding is consistent with the study by Gifford et al~\cite{gifford2020fatty}.

In case study 2, we employed DRExplainer and ExplaiNE to analyze the top five sensitive and resistant interactions according to the value of f1@k. We compared their outputs with our ground truth dataset to evaluate accuracy and relevance. The explanations, structured as triplets of cell lines, responses, and drugs, were assessed for their alignment with the ground truth. The results in Table~\ref{tab5} indicated that both DRExplainer and ExplaiNE exhibit competent explanatory power for resistant drug response. However, ExplaiNE showed limitations in explaining sensitive responses. For example, while DRExplainer consistently provided credible explanations for the sensitivity of the SKOV3 cell line to 5-FU, ExplainNE failed to find any consistent explanation.

Therefore, the two case studies demonstrate that DRExplainer could help find out the novel drug response and reveal its underlying mechanism. By aligning predictions with existing biological knowledge, the model highlighted its potential to guide personalized treatment strategies. Moreover, DRExplainer offered critical insights that can assist clinical decision-making and avoid ineffective therapies. These examples illustrate how the predictions of the model can directly influence real-world clinical practices and improve patient outcomes.

\begin{table*}[!b]
\caption{Top 10 predicted sensitive and resistant drug responses.\label{tab4}}

\begin{tabular*}{\textwidth}
{@{\extracolsep\fill}lllll@{\extracolsep\fill}}
\toprule
Response & Rank & Cell line & Drug & Evidence(PMID) \\
\midrule
\multirow{10}{*}{Sensitive} & 1 & OCIAML2 & 5-Fluorouracil & 25667168 \\
    & 2 & DND41 & 5-Fluorouracil & NA \\
    & 3 & OCILY19 & Ispinesib Mesylate & NA \\
    & 4 & BV173 & Omipalisib & NA \\
    & 5 & RPMI8402 & 5-Fluorouracil & 3873586 \\
    & 6 & BV173 & HG6-64-1 & NA \\
    & 7 & CMK & 5-Fluorouracil & 19458058 \\
    & 8 & RCHACV & Ispinesib Mesylate & NA \\
    & 9 & JHH4 & 5-Fluorouracil & NA \\
    & 10 & SKOV3 & 5-Fluorouracil & 22502731 \\
\midrule
\multirow{10}{*}{Resistant} & 1 & OCILY19 & DMOG & NA \\
\multicolumn{1}{c}{} & 2 & OCILY19 & Belinostat  & NA \\
\multicolumn{1}{c}{} & 3 & KASUMI1 & ACY-1215 & NA \\
\multicolumn{1}{c}{} & 4 & OCILY19 & NSC-207895 & NA \\
\multicolumn{1}{c}{} & 5 & NCIH1355 & LFM-A13 & NA \\
\multicolumn{1}{c}{} & 6 & NCIH1299 & Tubastatin A & NA \\
\multicolumn{1}{c}{} & 7 & OCILY19 & Etoposide & 32249639 \\
\multicolumn{1}{c}{} & 8 & OCILY19 & Idelalisib & 32723130 \\
\multicolumn{1}{c}{} & 9 & CORL311 & Tubastatin A & NA \\
\multicolumn{1}{c}{} & 10 & MDAMB231 & Bleomycin & 27006451 \\
\bottomrule
\end{tabular*}
\end{table*}

\begin{table*}[!t]
\caption{Case studies for the explanations of DRExplainer and ExplaiNE.\label{tab5}}
\centering
\renewcommand{\arraystretch}{0.8}
\footnotesize

\begin{tabular*}{\textwidth}{@{\extracolsep{\fill}}lllllp{8cm}@{\extracolsep{\fill}}}
\toprule%
Method & Response & Rank & Drug & Cell line & Matched explanation \\
\midrule
\multirow{28}{*}{DRExplainer} & \multirow{13}{*}{Resistant} & 1 & Tenovin-6 & HPAC & \footnote{}5 matched {[}(HPAC, \textit{3}, HT1197), (HPAC, \textit{3}, KU1919), (HPAC, \textit{0}, CX-5461), (Tenovin-6, \textit{2}, AS605240), (SNUC1, \textit{0}, Tenovin-6){]} \\
& & 2 & QL-XI-92 & NCIH841 & 5 matched {[}(NUGC4, \textit{0}, QL-XI-92), (CAL120, \textit{0}, QL-XI-92),   (RERFGC1B, \textit{0}, QL-XI-92), (NCIH2087, \textit{0}, QL-XI-92), (HCC1937, \textit{0}, QL-XI-92){]} \\
& & 3 & BIX02189 & SW48 & 5 matched {[}(SW48, \textit{3}, CAMA1), (NCIH1975, \textit{0}, BIX02189), (HCC1428, \textit{0}, BIX02189), (SW48, \textit{3}, CFPAC1), (SKUT1, \textit{3}, SW48){]} \\
& & 4 & Bleomycin & SW48 & 5 matched {[}(SW48, \textit{3}, CAMA1), (Bleomycin, \textit{0}, IALM), (SW48, \textit{3}, CFPAC1), (Bleomycin, \textit{0}, CAL29), (TL-2-105, \textit{2}, Bleomycin){]} \\                                
& & 5 & QS11 & SW48 & 5 matched {[}(SW48, \textit{3}, CAMA1), (ES2, \textit{0}, QS11), (SW48, \textit{3}, CFPAC1), (QS11, \textit{0}, EFM19), (CCK81, \textit{0}, QS11){]} \\
\cline{2-6}
\rule{0pt}{3ex}
& \multirow{15}{*}{Sensitive} & 1 & 5-Fluorouracil & SKOV3 & 5 matched {[}(OE33, \textit{1}, 5-Fluorouracil), (ONS76, \textit{1}, 5-Fluorouracil), (BL41, \textit{1}, 5-Fluorouracil), (SCC4, \textit{1}, 5-Fluorouracil), (GCT, \textit{3}, SKOV3){]} \\
& & 2 & 5-Fluorouracil & JIMT1 & 5 matched {[}(JIMT1, \textit{3}, OVCAR5),   (G401, \textit{1}, 5-Fluorouracil), (NCIH2228, \textit{1}, 5-Fluorouracil), (JHH6, \textit{1}, 5-Fluorouracil), (MFE319, \textit{1}, 5-Fluorouracil){]} \\
& & 3 & 5-Fluorouracil & DKMG & 5 matched {[}(DKMG, \textit{3}, TT2609C02), (DKMG, \textit{3}, NCIH358), (NCIH661, \textit{1}, 5-Fluorouracil), (EPLC272H, \textit{1}, 5-Fluorouracil), (BCPAP, \textit{1}, 5-Fluorouracil){]} \\
& & 4 & 5-Fluorouracil & OV90 & 5 matched {[}(OV90, \textit{3}, SKOV3), (NCIH661, \textit{1}, 5-Fluorouracil), (KYSE410, \textit{3} OV90), (EPLC272H, \textit{1}, 5-Fluorouracil), (BCPAP, \textit{1}, 5-Fluorouracil){]} \\
& & 5 & 5-Fluorouracil & KYSE150 & 5 matched {[}(G401, \textit{1}, 5-Fluorouracil), (NCIH2228, \textit{1}, 5-Fluorouracil), (JHH6, \textit{1}, 5-Fluorouracil), (MFE319, \textit{1}, 5-Fluorouracil), (KYSE150, \textit{3}, TT2609C02){]} \\
\midrule
\multirow{22}{*}{ExplaiNE} & \multirow{14}{*}{Resistant} & 1 & Tenovin-6 & HAPC & 5 matched {[}(Tenovin-6, \textit{2}, Omipalisib), (Tenovin-6, \textit{2}, PHA-793887), (Tenovin-6, \textit{2}, CGP-60474), (Tenovin-6, \textit{2}, Belinostat), (Tenovin-6, \textit{2}, QL-XII-47){]} \\                       
& & 2 & QL-XI-92 & NCIH841 & 4 matched {[}(NCIH841, \textit{0}, Idelalisib), (NCIH841, \textit{0}, SU11274), (NCIH841, \textit{0}, KIN001-270), (NCIH841, \textit{0}, CAP-232){]} + 1 unseen \\ 
& & 3 & BIX02189 & SW48 & 5 matched {[}(BIX02189, \textit{0}, TOV112D), (BIX02189, \textit{0}, BFTC905), (BIX02189, \textit{0}, JHH4), (BIX02189, \textit{0}, CAL29), (BIX02189, \textit{0}, SW837){]} \\
& & 4 & Bleomycin & SW48 & 5 matched {[}(Bleomycin, \textit{0}, UACC812), (Bleomycin, \textit{0}, ABC1), (Bleomycin, \textit{0}, T84), (Bleomycin, \textit{0}, BFTC905), (Bleomycin, \textit{0}, SW1417){]} \\
& & 5 & QS11  & SW48 & 5 matched {[}(QS11, \textit{0}, NCIH524), (QS11, \textit{2}, Embelin), (QS11, \textit{0}, COLO680N), (QS11, \textit{0}, SKMEL24), (QS11, \textit{0}, NCIH358){]} \\
\cline{2-6}
\rule{0pt}{3ex}
& \multirow{7}{*}{Sensitive} & 1 & 5-Fluorouracil & SKOV3 & 5 unseen \\
& & 2 & 5-Fluorouracil & JIMT1 & 2 matched {[}(JIMT1, \textit{3}, SNU449), (JIMT1, \textit{3}, MKN1){]} + 3 unseen \\
& & 3 & 5-Fluorouracil & DKMG & 4 matched {[}(DKMG, \textit{3}, KYSE150), (DKMG, \textit{3}, HCC1395), (DKMG, \textit{3}, PC14), (DKMG, \textit{3}, HT29){]} + 1 unseen \\
& & 4 & 5-Fluorouracil & ACHN & 5 unseen \\
& & 5 & 5-Fluorouracil & OV90 & 5 unseen \\  
\bottomrule
\end{tabular*}
\begin{tablenotes}%
\item Note: \textit{matched} indicates the explanations are generated and also in the ground truth dataset, while \textit{unseen} denotes the generated explanations are not in the ground truth. In the ``Matched explanation" column, the values \textit{0}, \textit{1}, \textit{2}, \textit{3} represent resistant response, sensitive response, cell line similarity and drug similarity, respectively.
\end{tablenotes}
\end{table*}

\section{Conclusion}
Nowadays, the heterogeneous graphs encompassing drugs and cell lines have emerged as a powerful tool for drug response prediction. The great potential of the GNN model based on a directed bipartite network has not been fully exploited. For precision medicine, stakeholders demand not only precise predictions but also comprehensive insights into the models' decision-making processes. In this work, we propose DRExplainer, an interpretable directed graph convolutional network for drug response prediction. DRExplainer leverages multi-source bio-entity information to boost model performance. Additionally, we curate a ground truth benchmark dataset specifically to assess the interpretability of the model. The experimental results and case studies demonstrate that DRExplainer is superior to existing state-of-the-art methods in identifying and explaining novel drug responses.

The implications of this study extend from theoretical advancements in drug response prediction to practical benefits for clinical and preclinical applications. First, by providing interpretable predictions, DRExplainer offers reliable insights to support clinical decision-making and presents a novel approach to advancing the clinical adoption of deep learning models in oncology. Second, the ability of DRExplainer to predict both sensitivity and resistance response can streamline the preclinical evaluation of potential therapies, reducing the time and cost associated with developing new drugs. Lastly, this work advances precision medicine by mitigating the trial-and-error approach in cancer treatment, enabling more patients to receive effective, personalized therapies tailored to their unique profiles.

In future work, we provide two directions to enhance interpretable drug response prediction: (i) Collect single-cell omics data and apply them to predict drug responses. Single-cell omics offer an unparalleled resolution, enabling the exploration of cellular heterogeneity that often remains concealed with bulk omics techniques. Leveraging these insights, future studies will aim to integrate single-cell data into our directed graph convolutional network model, enhancing its predictive accuracy and interoperability. (ii) Enrich our model by incorporating extensive domain knowledge into constructing ground truth datasets. By integrating comprehensive insights from biological research, such as pathways, gene regulation mechanisms, and cellular interactions, we can significantly improve the fidelity of our predictive models to actual biological processes. This enriched ground truth framework will facilitate the development of a paradigm model in biological issue prediction, establishing benchmarks that blend high predictive accuracy with deep biological understanding. Such advancements are expected to set new standards in the field, propelling forward the capabilities of computational tools in precision medicine and beyond.

\section{CRediT authorship contribution statement}
\textbf{Haoyuan Shi:} Writing - original draft, Writing - review \& editing, Methodology, Formal analysis, Visualization. \textbf{Tao Xu:} Investigation, Data curation, Validation, Visualization. \textbf{Xiaodi Li:} Data curation, Validation, Visualization. \textbf{Qian Gao:} Data curation, Validation, Visualization. \textbf{Zhiwei Xiong:} Formal analysis, Writing - review \& editing.  \textbf{Junfeng Xia:} Supervision. \textbf{Zhenyu Yue:} Supervision, Project administration, Writing - review \& editing.

\section{Declaration of competing interest}
The authors declare that they have no known competing financial interests or personal relationships that could have appeared to influence the work reported in this paper.

\section{Acknowledgements}
This work was supported by the grants from the National Natural Science Foundation of China (62472005, 62102004), and the Anhui Province Excellent Young Teacher Training Project (YQYB2024007).


\end{document}